\documentclass[a4paper]{article}

\usepackage{INTERSPEECH2019}
\usepackage[toc,page]{appendix}
\usepackage{url,verbatim}
\usepackage[utf8]{inputenc}
\usepackage{CJKutf8}

\newcommand\blankfootnote[1]{%
  \let\thefootnote\relax\footnotetext{#1}%
  \let\thefootnote\svthefootnote%
}

\title{CSS10: A \underline{C}ollection of \underline{S}ingle Speaker \underline{S}peech Datasets for \underline{10} Languages}
\name{Kyubyong Park$^{*, \dagger}$, Thomas Mulc$^{\wedge,\dagger}$%\\
}
\address{
  $^{*}$Kakao Brain, South Korea\\
  $^{\wedge }$Expedia Group, U.S.A.}
\email{kyubyong.park@kakaobrain.com, tmulc@expedia.com}

\begin{document}

\maketitle
\begin{abstract}
  We describe our development of CSS10, a collection of single speaker speech datasets for ten languages. It is composed of short audio clips from LibriVox audiobooks and their aligned texts. To validate its quality we train two neural text-to-speech models on each dataset. Subsequently, we conduct Mean Opinion Score tests on the synthesized speech samples. We make our datasets, pre-trained models, and test resources publicly available. We hope they will be used for future speech tasks.

\end{abstract}

\blankfootnote{$^{\dagger}$ Both authors contributed equally.}

\section{Introduction}
Text-to-speech (TTS) models aim at generating an audio sequence given some input text. Recently there have been many TTS models based on neural networks. Among them are WaveNet \cite{wavenet}, Tacotron 1 \& 2 \cite{tacotron}\cite{tacotron2}, Char2Wav \cite{char2wav}, DeepVoice 1, 2, \& 3 \cite{dv1}\cite{dv2}\cite{dv3}, DCTTS \cite{dctts}, VoiceLoop \cite{voiceloop}, etc. All of them adopted an end-to-end approach, taking as input a sequence of characters or phonemes and returning the raw waveform or spectrogram. 
\\
 \indent All of those models have been introduced with results manufactured on their internal datasets except DCTTS, which uses the public LJ Speech Dataset \cite{ljspeech17}. This status quo has made reproducing results in the papers difficult for researchers outside. People have resorted to using less than ideal data and/or creating their own datasets that try to match some of the properties of the internal ones. What is even worse is that it is hard to compare different TTS models, because there is no benchmark dataset. 
\\
\indent Furthermore, the research is mostly focused on English. Except Deep Voice 2 \& 3, which trained on the Chinese language, all the other models attempted to model the text-to-audio mapping for the English language. The most likely explanation is simply the dearth of freely available non-English datasets. We strongly believe that languages other than English deserve to have attention from researchers as well. This is why we construct multi-lingual speech datasets and share them with the research community.
\\
\indent Our contribution is two-fold:
\begin{itemize}
\item We construct single speaker speech datasets with aligned text for ten different languages.
\item We train two famous TTS models on each dataset and evaluate them with Mean Opinion Scores (MOS).
\end{itemize}
All of the resources mentioned above are available in our GitHub repository$^1$\blankfootnote{$^1$ \url{https://github.com/Kyubyong/CSS10}}.

\section{Related work}
Not surprisingly, for the English language, there are many public speech datasets such as Blizzard \cite{blizzard}, VCTK \cite{vctk}, LibriSpeech \cite{librispeech}, TED-LIUM \cite{tedlium}, VoxForge \cite{voxforge}, and Common Voice \cite{commonvoice}. While these datasets are large---some having more than 100 hours of audio---they are all from multiple speakers, which make them unideal for the single speaker TTS task (namely, generating speech from text in the voice of a single speaker) where more data from a single speaker tends to help model performance \cite{cmu_org}. One popular, newly-created dataset is the LJ Speech dataset \cite{ljspeech17}. It consists of audio files segmented from audiobooks in LibriVox recorded by a female volunteer. It has a large cumulative audio length (20+ hrs) and has been verified to work with neural TTS models. Another is the World English Bible (WEB) dataset \cite{web}, which was sourced from bible recordings and text in the public domain. It shares the LJ Speech dataset's properties but is sampled at a relatively low rate of 12 kHz. Both are publicly available.
\\
\indent For the Japanese language, there is a public dataset called the JSUT dataset \cite{jsut}. It is a single speaker dataset designed for speech synthesis that includes around 10 hours of utterances with aligned text. One important distinction between the JSUT dataset and the LJ Speech and WEB datasets is that the JSUT dataset's recording was carried out in a controlled environment with specially designed scripts. One drawback of the JSUT dataset is that it lacks a phonetic transcription of text. Additionally, there were no follow-up experiments on the dataset. 
\\
\indent For the German language, there is the Pavoque dataset \cite{pavoque}---a single speaker, multi-style corpus of German speech. It has 12+ hours of audio clips each of which is associated with phoneme-level annotations. However, it is more suitable for speech style tasks rather than regular TTS tasks.
\\
\indent  A dataset closer to our work is the Spoken WP Corpus Collection \cite{spokenwp}. It has hundreds of hours of audio with aligned text for three languages: Dutch, German, and English. For each language, much of the audio comes from a single speaker. However, the audio clips are generally large---on the order of minutes---which makes using the dataset difficult for neural TTS models \cite{tundra}.
\\
\indent Perhaps the datasets closest to our work are the Tundra \cite{tundra} and M-AILABS \cite{mailabs} datasets, which have 14 and 9 languages, respectively, and are built from audiobooks. The Tundra dataset uses a single speaker for each language, but does so by using only one audiobook per language. The M-AILABS dataset, which focuses on European languages, has nearly one-thousand hours of total audio but uses multiple speakers in each language.
\section{Datasets}

\indent We choose to use audiobooks from LibriVox \cite{librivox}, a website for free public domain audiobooks, as the source of audio data for three major reasons. First, audiobooks are inherently accompanied by text, which is essential for our purpose. Second, although many readings are performances and hence use abnormal intonation, readers tend to speak with a regular, constant speed. Third, many audiobooks are performed by a single speaker.

\subsection{Selection of audiobooks}
As of now, there are audiobooks for 95 languages in LibriVox. We examine how many hours of solo recordings each language has; many audiobook performers record multiple audiobooks which makes finding large quantities of single speaker data achievable. We exclude a language if it has less than 4 hours of solo audio recordings, because we are not confident that training will succeed$^2$\blankfootnote{$^2$This number was chosen based on our prior successes of training DCTTS on an audiobook with 4 hours of data.}. Then, we check for text availability and for audio quality. If text is not available or if the audio includes a noticeable amount of noise, the audiobook is excluded.
\\
\indent This process yields audiobooks for the following languages: Chinese (zh), Dutch (nl), French (fr), Finnish (fi), German (de), Greek (el), Hungarian (hu), Japanese (ja), Russian (ru), and Spanish (es)$^3$\blankfootnote{$^3$The parenthetical expression next to each language is the language code from ISO 639-1.}. All audio files are sampled at 22 kHz. 

\subsection{Audio processing}
\begin{figure}[t]
  \centering
  \includegraphics[width=\linewidth]{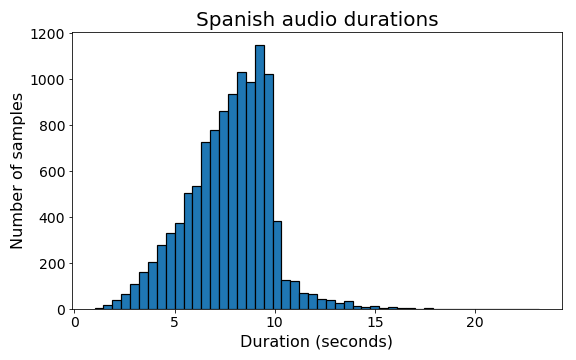}
  \caption{Histogram of audio durations for the Spanish dataset.}
  \label{fig:dist}
\end{figure}

The audio from LibriVox usually comes in large files with lengthy audio clips that do not suit the TTS task, so we fragment them into many small files. We use the audio editor Audacity \cite{audacity} to programmatically find split-points of the audio anytime there is more than a 0.5-second duration of silence, except for Spanish audiobooks, where we use a 0.25-second duration. Next, we adjust the points such that neighboring clips are joined to have a duration around 10 seconds. We found these tricks improve computational efficiency. The distribution of audio lengths for the Spanish dataset is shown in Figure \ref{fig:dist}. We see about 85\% of the samples have a duration between 5 and 11 seconds. All other languages have distributions similar to this.

\subsection{Text processing}
We have experts align the text with each segmented audio clip to create an $<$audio, text$>$ pair. At first we considered using a forced aligner such as Gentle \cite{gentle}. However, we forfeited the idea after we realized that it does not guarantee the correct alignments and that it is language dependent.

\subsubsection{Text normalization}
Once we secure the $<$audio, text$>$ pairs, we request that our experts normalize text; all abbreviations are expanded (e.g., Dra. $\rightarrow$ Doctora) and Arabic numerals are spelled out to match the context. Unlike Deep Voice 3 or DCTTS, which ignore case, we decide to retain case, because it can be a cue for new sentences or proper nouns or nouns. In addition, we remove infrequent symbols other than punctuations such as the period, question mark, exclamation point, colon, comma, and semi-colon. 

\subsubsection{Phonetic transcription}

\indent Because we take text as input, it is important to understand the writing system of our target languages. Dutch, German, Finnish, French, Hungarian, and Spanish are based on Latin alphabets, while Greek and Russian use Greek letters and Cyrillic letters, respectively. Chinese uses Chinese characters, and Japanese employs three different scripts (i.e., Hiragana, Katakana, and Chinese characters). All the writing systems except Chinese characters are phonetic. In other words, they are written as they are pronounced. On the other hand, Chinese characters are ideographic, so they are not directly associated with pronunciations. Thus, Chinese people typically use the Romanization system for pronunciation, which is called pinyin, to input Chinese characters in digital settings. Therefore, we convert original Chinese text into pinyin (e.g., \begin{CJK}{UTF8}{mj} 思想言论举动丰采都没有什么大区别 \end{CJK} $\rightarrow$ sī xiǎng yán lùn jǔ dòng fēng cǎi dū méi yǒu shén me dài qū bié) using the Chinese segmenter Jieba \cite{jieba} and the open-source dictionary CC-CEDICT \cite{cedict}. For Japanese, first we use a morphological analyzer MeCab \cite{mecab} to get the pronunciations of the text and subsequently use romkan \cite{romkan} to convert them into Roman letters. When MeCab fails to return the pronunciations for words, we have a native speaker create them manually.
\\
\indent Now, each of our datasets contains $<$audio, original text, processed text$>$ triplets.

\section{Experiments}
In order to validate our datasets, we train two neural TTS models---Tacotron \cite{tacotron} and DCTTS \cite{dctts}---on each dataset. We then synthesize audio samples and perform Mean Opinion Score tests.

\subsection{Models}
Tacotron and DCTTS are well-known neural TTS models that have an attention-based sequence-to-sequence architecture. Tacotron, which was introduced in 2017, is an impactful model that produced high quality results from an end-to-end approach and has since served as an important benchmark. DCTTS is one of the TTS models inspired by Tacotron. DCTTS is different from Tacotron in a few ways. Tacotron's computational backbone is the CBHG, a combination of convolution layers, fully-connected layers, and recurrent layers, whereas DCTTS only uses convolution layers. Tacotron is trained end-to-end whereas DCTTS has two networks trained independently. DCTTS uses a few tricks that help training such as guided attention and forcibly incremental attention.  
\\
\indent One reason for choosing these models is that we have working implementations of both \cite{tacotron_imp}\cite{dctts_imp} which have been tested on multiple datasets. We have found that reproducing original work in neural TTS is non-trivial. Model performance is dependent on many factors, such as data, model architecture, hyperparameters, etc. The neural TTS field has yet to have the ossification of helpful techniques for training that has seemingly already happened in the neighboring field of neural computer vision. Thus, we find it paramount to use models that have successful implementations in order to evaluate new datasets.
\subsection{Training}
We use Tesla P40 GPUs for training. For both models, we mostly use the given hyperparameters in their respective papers. See our repository for the full list of values. %$^4$\blankfootnote{$^4$\url{https://github.com/Kyubyong/css10/blob/master/tacotron/hyperparams.py}} and DCTTS$^5$\blankfootnote{$^5$\url{https://github.com/Kyubyong/css10/blob/master/dc_tts/hyperparams.py}}. 
We use the same hyperparameters for all languages. The original Tacotron paper uses more than 2 million training steps, which we find impractical given our resources. In our preliminary tests, we found 400k steps produced good results for DCTTS. For simplicity, we train both models for 400k steps. It takes around ten days and three days to train Tacotron and DCTTS, respectively.

\subsection{Evaluation}
\subsubsection{Test sentences}
To evaluate each model's performance, 20 sentences per language are collected from Tatoeba \cite{tatoeba}, a web database of sentences for multiple languages. These sentences are carefully chosen to maximize the cover of letters in the vocabulary so we can check the utterance quality of various phonemes. For Chinese and Japanese, phonetic transcriptions, not the original text, are considered. Some letters which are very rare in the language are left out. All sentences are available in our repository.

% We release all evaluation sentences to the public$^7$ \blankfootnote{$^7$\url{https://github.com/Kyubyong/css10/tree/master/MOS/sents}}.
\subsubsection{Mean Opinion Scores}
We leverage Amazon Mechanical Turk (MTurk) to gather workers to score the test sentences. MTurk allows requesters to post Human Intelligence Tasks (HITs) for a worker to complete. For each HIT, we ask the worker to listen to an audio clip and score it.
\\
\indent
We adopt the standard absolute category rating (ACR) test, where workers are required to give integer scores between 1 and 5. We split evaluation into two categories: speech naturalness and pronunciation accuracy. For speech naturalness, we refer to the rubric for MOS scores in \cite{crowdmos}, which we include in Table \ref{tab:nat_acc} for completeness. We design a new simple rubric for pronunciation accuracy as shown in Table \ref{tab:pron_acc}.

\begin{comment}
\begin{table*}[th]
  \caption{Rubrics of naturalness and pronunciation accuracy.}
  \label{tab:nat_acc}
  \centering
  \begin{tabular}{p{.5cm} p{1cm}  p{6cm} p{4cm}}
    \toprule
    \multicolumn{1}{c}{\textbf{Score}} &                            
    \multicolumn{1}{c}{\textbf{Quality}} &
    \multicolumn{1}{c}{\textbf{Naturalness Definition}} &
    \multicolumn{1}{c}{\textbf{Pronunciation Accuracy Definition}}\\
    \midrule
    5 &  Excellent & Imperceptible distortions & No mispronunciations \\
    4 &  Good & Just perceptible but not annoying distortions & Few minor mispronunciations \\
    3 &  Fair & Perceptible and slightly annoying distortions & Many minor mispronunciations \\
    2 &  Poor & Annoying but not objectionable distortions & Few major mispronunciations \\
    1 &  Bad & Very annoying and objectionable distortions & Many major mispronunciations \\
    \bottomrule
  \end{tabular}
\end{table*}
\end{comment}

%\begin{comment}
\begin{table}[th]
  \caption{Rubric of naturalness.}
  \label{tab:nat_acc}
  \centering
  \begin{tabular}{p{.1cm} p{1.2cm}  p{5cm}}
    \toprule
    \multicolumn{1}{c}{\textbf{Score}} &                            
    \multicolumn{1}{c}{\textbf{Quality}} &
    \multicolumn{1}{c}{\textbf{Definition}}\\
    \midrule
    5 &  Excellent & Imperceptible distortions (dist.)\\
    4 &  Good & Just perceptible but not annoying dist.\\
    3 &  Fair & Perceptible and slightly annoying dist.\\
    2 &  Poor & Annoying but not objectionable dist.\\
    1 &  Bad & Very annoying and objectionable dist.\\
    \bottomrule
  \end{tabular}
\end{table}

\begin{table}[th]
  \caption{Rubric of pronunciation accuracy.}
  \label{tab:pron_acc}
  \centering
  \begin{tabular}{p{.5cm} p{1.2cm}  p{5cm}}
    \toprule
    \multicolumn{1}{c}{\textbf{Score}} &                            
    \multicolumn{1}{c}{\textbf{Quality}} &
    \multicolumn{1}{c}{\textbf{Definition}}\\
    \midrule
    5 &  Excellent & No mispronunciations\\
    4 &  Good & Few minor mispronunciations\\
    3 &  Fair & Many minor mispronunciations \\
    2 &  Poor & Few major mispronunciations  \\
    1 &  Bad & Many major mispronunciations \\
    \bottomrule
  \end{tabular}
\end{table}
%\end{comment}

 With each HIT, we give the worker a reference sample from the audiobook and tell the worker that the sample should score highly on both naturalness and pronunciation; we then give the worker an audioclip with the corresponding text and ask them score it.
 
\indent For some languages, MTurk allows us to list qualifications for basic proficiency. While we could opt for this requirement for some languages (e.g., French), we want consistency in the MOS procedure across all languages to make the results more comparable. Thus, instead of relying on it, we require each worker to listen to a reference sample chosen from the language's dataset and transcribe the text. We consider participants who have done this correctly to be truthful and include their scores in our results.
\\
\indent
We use the same method as crowdMOS \cite{crowdmos} for computing confidence intervals (C.I.). Because the number of available workers to complete HITs varies with language, we allow languages to have varying numbers of total samples.

\subsection{Results}
\begin{figure}[t]
  \centering
  \includegraphics[width=\linewidth]{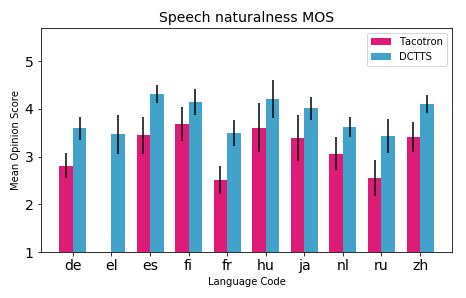}
  \caption{MOS scores for speech naturalness with 95\% C.I.}
  \label{fig:bar audio}
\end{figure}

\begin{figure}[h!]
  \centering
  \includegraphics[width=\linewidth]{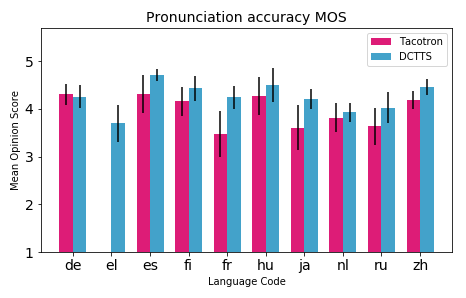}
  \caption{MOS scores for pronunciation accuracy with 95\% C.I.}
  \label{fig:bar audio2}
\end{figure}

We successfully trained models for all languages except Greek for Tacotron. We believe it is due to the small data size of Greek ($\sim$4 hours of audio). Although we successfully trained DCTTS on Greek, the samples were much worse than those from other languages.
\\ 
\indent The MOS results for speech naturalness and pronunciation accuracy are shown in Figures \ref{fig:bar audio} and \ref{fig:bar audio2}, respectively. For naturalness, DCTTS is statistically more natural than Tacotron for German, French, and Spanish. For pronunciation accuracy, we find that DCTTS and Tacotron are somewhat similar. The detailed MOS scores are found in Table \ref{tab:datasets}.

\begin{table*}[th]
  \caption{MOS scores with 95\% C.I. of Tacotron and DCTTS for all languages.}
  \label{tab:datasets}
  \centering
  \begin{tabular}{ l l l l l l l l l l}
    \toprule
    \textbf{Lang.} &                            
    \textbf{Dur. (hh:mm:ss)} &
    \multicolumn{1}{c}{\textbf{\# Workers}} &
    \multicolumn{2}{c}{\textbf{Speech Naturalness}} &
    \multicolumn{2}{c}{\textbf{Pronunciation Accuracy}}\\
    & &  & Tacotron & DCTTS
    & Tacotron & DCTTS\\
    \midrule

    de &  16:08:01 & 75 & $2.82 \pm 0.26$ & $3.59 \pm 0.24$ & $4.31 \pm 0.22$ & $4.25 \pm 0.24$\\
    el &  04:08:14 & 43   & N/A & $3.47 \pm 0.41$ & N/A & $3.70 \pm 0.39$\\
    es &  23:49:49 & 78 & $3.45 \pm 0.38$ & $4.31 \pm 0.19$ & $4.31 \pm 0.40$ & $ 4.72 \pm 0.13$\\
    fi &  10:32:03 & 62 & $3.69 \pm 0.36$ & $4.15 \pm 0.27$ & $4.16 \pm 0.30$ & $ 4.43 \pm 0.27$\\
    fr &  19:09:03 & 75  & $2.51 \pm 0.30$ & $3.50 \pm 0.27$ & $3.48 \pm 0.48$ & $4.24 \pm 0.25$\\
    hu &  10:00:25 & 39 & $3.60 \pm 0.51$ & $4.21 \pm 0.39$ & $4.27 \pm 0.40$ & $ 4.51 \pm 0.35$\\
    ja &  14:55:36 & 79 & $3.39 \pm 0.48$ & $4.01 \pm 0.24$ & $3.61 \pm 0.48$ & $4.20 \pm 0.21$\\
    nl &  14:06:40 & 97  & $3.06 \pm 0.35$ & $3.63 \pm 0.21$ & $3.82 \pm 0.31$ & $3.93 \pm 0.20$\\
    ru &  21:22:10 & 73 & $2.55 \pm 0.37$ & $3.43 \pm 0.35$ & $3.64 \pm 0.39$ & $4.02 \pm 0.32$\\
    zh &  06:27:04 & 122 & $3.41 \pm 0.31$ & $4.10 \pm 0.19$ & $4.19 \pm 0.20$ & $4.46 \pm 0.17$\\
    \bottomrule
  \end{tabular}
\end{table*}

\begin{figure}[t]
  \centering
  \includegraphics[width=\linewidth]{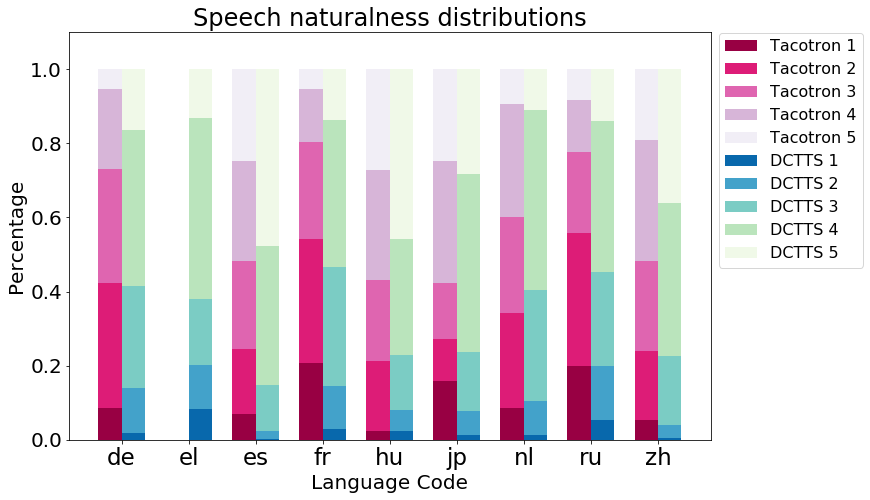}
  \caption{Percentages of naturalness scores shown by model and language.}
  \label{fig:per nat}
\end{figure}

\begin{figure}[h!]
  \centering
  \includegraphics[width=\linewidth]{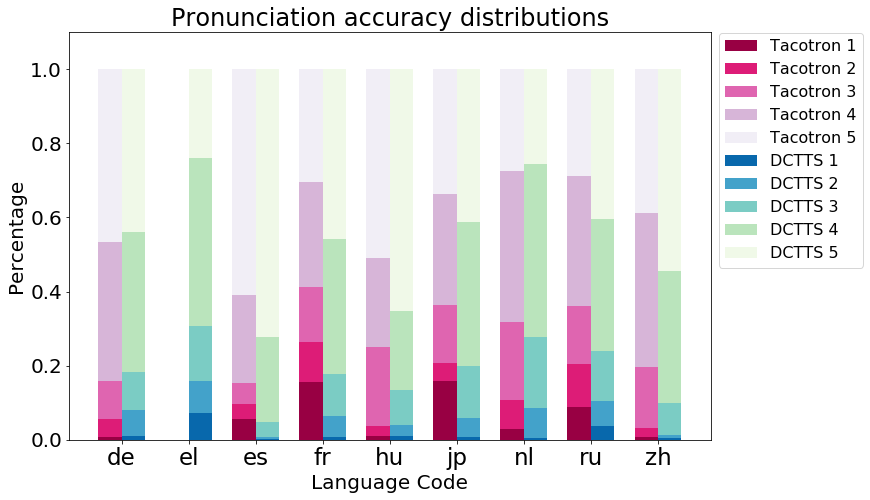}
  \caption{Percentages of pronunciation accuracy scores shown by model and language.}
  \label{fig:per acc}
\end{figure}

Although it's common to use MOS as a performance metric for TTS models, we recognize that it may not be appropriate to take the mean of Likert scores, because each score belongs to a category and the semantic meaning of these categories need not be evenly distributed along a number line as we (and others) have implied in our rubrics. Thus, we also present the distributions of scores for each model-language pair; the distributions for naturalness and pronunciation accuracy are shown in Tables \ref{fig:per nat} and \ref{fig:per acc}, respectively. We see that for both speech naturalness and pronunciation accuracy, DCTTS has distributions more skewed towards higher scores than Tacotron, but the effect is more marked for naturalness.% pronunciation accuracy DCTTS and Tacotron have similar distributions.

%We see that for both naturalness and pronunciation, DCTTS has a higher percentage of the top score (5) for almost all languages. Additionally, for naturalness, DCTTS has a distribution more skewed to higher scores than Tacotron.

In general, when listening to samples, both models capture the original voice, but the samples of DCTTS sound better than those of Tacotron. We found DCTTS produced fairly clean speech, while Tacotron consistently had noise in the outputs.  
\\
\indent Both models exhibited mumbling at the end of a few generated samples. Although this was not expected--the models should have learned trailing silences--we were able to remove mumbling via trimming.
\\
\indent In Japanese, we found utterances that were not relevant to the input text were generated at the end in a few samples. We suspect it is derived from the imperfection of the automatic phonetic transcription in the training data.

\section{Conclusions}
We discussed how we built CSS10, a collection of single-speaker speech datasets for 10 languages, and how we used it for a TTS task. Despite the fact that there were differences in model performance depending on the language, we were able to train the models successfully on our datasets. We release all the resources for this project including source code, datasets, pre-trained models, and evaluation data to the public.

\section{Future work}
We hope CSS10 and our experiments serve as benchmarks for future non-English TTS research. 
%Particularly, we expect further investigation on the Japanese and Chinese languages. 
We found that some automatic phonetic transcriptions for Japanese and Chinese contain errors. If these errors are corrected, perhaps the model performance will improve. Additionally, we plan to add a Korean dataset. Because there are not enough audiobooks available for Korean in LibriVox, we are willing to produce recordings ourselves. When they are ready, we will release them with our other languages. Although we validated our datasets on TTS models in this work, CSS10 can be used for other speech tasks such as multi-lingual speech recognition.

\section{Acknowledgments}
We thank the LibriVox creator for his platform which allows public access to audiobooks, the performers who recorded their readings of audiobooks, and Kakao Brain for funding this work.%the %computational resources, the participants for text alignments, and the crowdsourcing.

%\clearpage

\bibliographystyle{IEEEtran}
%\newpage
\bibliography{paperbib}

\end{document}